\documentclass[lettersize,journal,revirew]{IEEEtran}
\usepackage{amsmath,amsfonts}
\usepackage{algorithmic}
\usepackage{algorithm}
\usepackage{array}
\usepackage[caption=false,font=normalsize,labelfont=sf,textfont=sf]{subfig}
\usepackage{textcomp}
\usepackage{stfloats}
\usepackage{url}
\usepackage{verbatim}
\usepackage{graphicx}
\usepackage{booktabs}
\usepackage{graphicx}
\usepackage{cite}
\usepackage{booktabs}
\usepackage{multirow}
\usepackage{graphicx}
\begin{document}
	
	\title{Visual Anomaly Detection Via Partition Memory Bank Module
		and Error Estimation}

	\author{Peng~Xing,
		Zechao~Li
		
		\IEEEcompsocitemizethanks{
			\IEEEcompsocthanksitem P. Xing, Z. li are with the School of Computer Science and Engineering, Nanjing University of Science and Technology, Nanjing 21094, China. E-mail: xingp\_ng@njust.edu.cn, zechao.li@njust.edu.cn. (Corresponding Author: Zechao Li)}
	}

	
	\markboth{Submission for IEEE Transactions on Circuits and Systems for Video Technology}%
	{Shell \MakeLowercase{\textit{et al.}}: IEEE TRANSACTIONS ON IMAGE PROCESSING}
	
	
	\maketitle

\begin{abstract}
Reconstruction method based on the memory module for visual anomaly detection attempts to narrow the reconstruction error for normal samples while enlarging it for anomalous samples. Unfortunately, the existing memory module is not fully applicable to the anomaly detection task, and the reconstruction error of the anomaly samples remains small. Towards this end, this work proposes a new unsupervised visual anomaly detection method to jointly learn effective normal features and eliminate unfavorable reconstruction errors. Specifically, a novel Partition Memory Bank (PMB) module is proposed to effectively learn and store detailed features with semantic integrity of normal samples. It develops a new partition mechanism and a unique query generation method to preserve the context information and then improves the learning ability of the memory module. The proposed PMB and the skip connection are alternatively explored to make the reconstruction of abnormal samples worse. To obtain more precise anomaly localization results and solve the problem of cumulative reconstruction error, a novel Histogram Error Estimation module is proposed to adaptively eliminate the unfavorable errors by the histogram of the difference image. It improves the anomaly localization performance without increasing the cost. To evaluate the effectiveness of the proposed method for anomaly detection and localization, extensive experiments are conducted on three widely-used anomaly detection datasets. The encouraging performance of the proposed method compared to the recent approaches based on the memory module demonstrates its superiority.
\end{abstract}


\section{Introduction}
\IEEEPARstart{V}{isual} anomaly detection aims to detect abnormal data that are different from normal visual data \cite{DBLP:conf/nips/GolanE18,zimek2012survey,leung2005unsupervised}, which has shown great potential in a variety of applications, such as industrial anomaly detection  \cite{bergmann2019mvtec,mo6587741,bergmann2020uninformed,mei2018automatic,9701300} and medical diagnosis \cite{li2018thoracic}. Since the abnormal patterns are diverse and the occurrence frequency of visual anomaly data is much lower than the normal ones, it is impractical to obtain sufficient abnormal training samples with different abnormal patterns. Consequently, it is challenging to successfully detect the anomaly data by using only normal training data.

\begin{figure}
	\centering
	\includegraphics[width=1\linewidth]{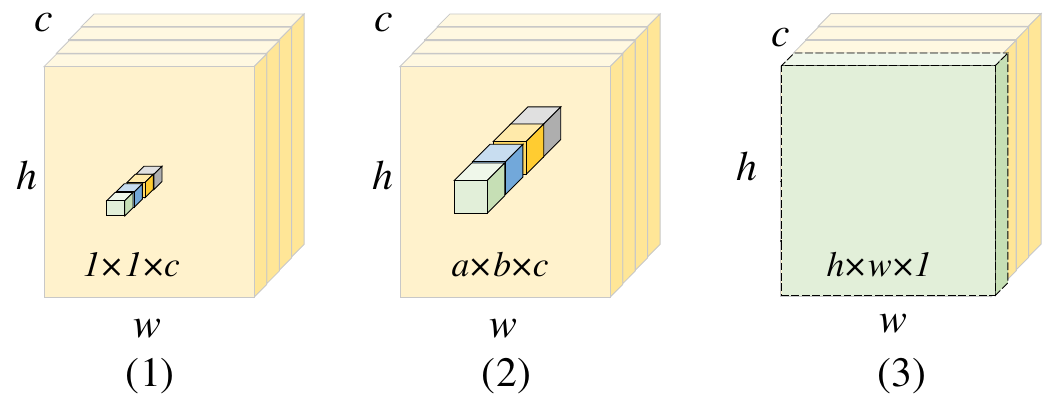}
	\caption{Illustration of the query generation methods for different memory modules. Figures (1) \cite{gong2019memorizing} and (2) \cite{hou2021divide} represent the logical query generated from the same position in the feature map, with different channels.  Figure (3) represents the proposed query. It preserves the critical features of original feature map learned by the convolutional network.}
	\label{fig:tuhw1}
\end{figure}

\par
Some methods have been studied to address the anomaly detection task. Recently, in \cite{zhai2016deep, zong2018deep,ber2019}, 
Autoencoder (AE) \cite{kingma2013auto} is introduced for unsupervised anomaly detection to detect abnormal samples based on reconstruction errors \cite{wold1987principal,xiong2011group, zhai2016deep,salehi2020arae}. 
These methods expect AE to reconstruct normal samples with only small reconstruction errors and reconstruct abnormal samples with large reconstruction errors.
It is well known that AE has a strong reconstruction capability \cite{gong2019memorizing,park2020learning} in many computer vision tasks. \cite{kumar2016ask,fan2019heterogeneous,DBLP:journals/tcsv/LuZZZH21,DBLP:journals/tcsv/LiuSC21}. However, the powerful reconstruction ability is not beneficial for anomaly detection. Even for unlearned anomalous samples, AE can still reconstruct them, resulting in small reconstruction error for unlearned anomalous samples \cite{gong2019memorizing}. Therefore, the reconstruction error of AE reconstruction alone is not sufficient to solve the anomaly detection challenge. Some works propose more complex self-supervised tasks, e.g., image-colorization \cite{fei2020attribute}. The complex self-supervised task makes AE learn normal semantic information to recover the original image. For the unlearned abnormal samples, the abnormal semantics cannot be recovered, making the reconstruction error larger. However, with tiny differences in semantic information between abnormal and normal samples, these methods can not detect the abnormal samples quite well.
 \begin{figure}
 	\centering
 	\includegraphics[width=1\linewidth]{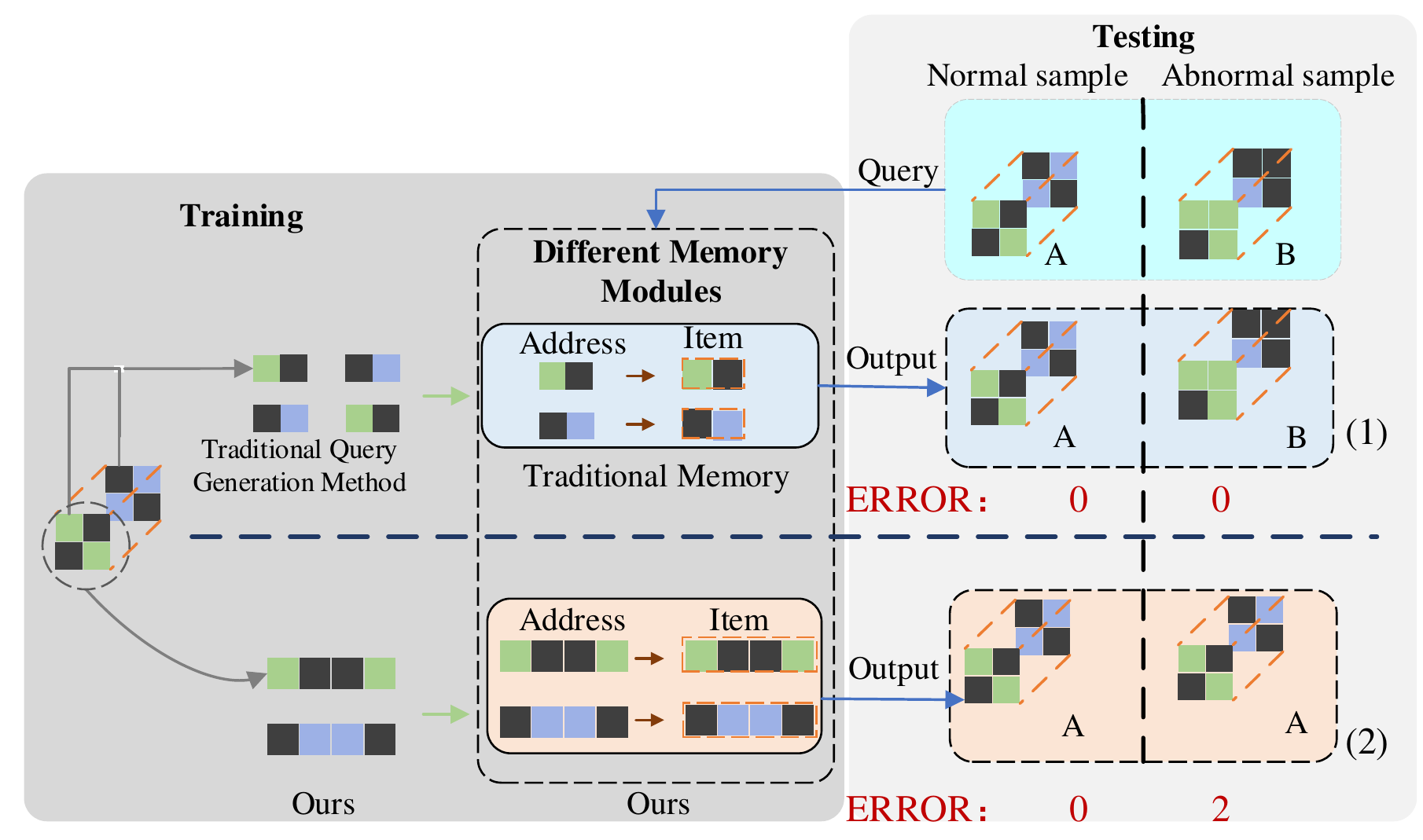}
 	\caption{The differences in query generation method and memory item between the proposed method and the traditional method. "ERROR" represents reconstruction error. The memory modules are both learned from the reconstruction task of category A to category A using reconstruction loss update. However, when the anomalous sample B is different from the normal sample A, the traditional methods can still successfully reconstruct B ("ERROR" is 0), while the proposed method fails to reconstruct unlearned category samples ("ERROR" is 2), which makes it successfully detect the anomalous sample B.}
 	\label{fig:fig1}
 \end{figure}

 To make the reconstruction errors of abnormal samples larger than the ones of normal samples, some methods \cite{gong2019memorizing,park2020learning,wang2021cognitive,hou2021divide} combine AE and the traditional memory modules. The output of the encoder is used as the query of a memory module, and the read of this memory module is used as the input of the decoder. Then, the memory module stores the normal features from the encoder output. When the abnormal feature queries the memory module, the output is normal feature. Therefore, the reconstruction error of abnormal sample becomes large. However, due to the query generation method, the existing memory modules do not actually store normal features, but rather logical information. As shown in Figure \ref{fig:tuhw1} (1) and Figure \ref{fig:tuhw1} (2), the query generated by the traditional method is a logical reorganization of features at the same location but from different channels of the original feature map \cite{park2020learning} \cite{hou2021divide}. A single channel feature is extracted by the same convolutional kernel by the sliding window method from the feature map of the previous layer, which contains similar critical feature information. Nevertheless, the queries generated by the logical reorganization of the feature map destroy the critical feature learned by the convolutional network, allowing the memory module to learn the reorganized logical information. 
 When anomalous features as query address memory module, the traditional module will fit the anomalous feature reads in logical relationship to support image reconstruction.
  Therefore, it makes the detection and localization of anomalous samples fail with small reconstruction errors. Figure \ref{fig:fig1} (1) shows an example of detection failure for anomalous sample B, which learns a memory module by the traditional query generation method (Figure \ref{fig:tuhw1} (1)).
 
 When facing tiny region anomaly samples, there is another aspect to consider - the effect of cumulative reconstruction error. Introducing the memory module makes it impossible for AE to achieve zero-error reconstruction even for normal regions. Due to the cumulative error, the abnormal scores of some normal samples can surpass the abnormal sample scores of small abnormal regions (e.g., small scratches), leading to incorrect detection and localization results. 
 
 Moreover, Existing methods \cite{gong2019memorizing,park2020learning} adopt a scheme, which concatenates the outputs of last encoder and the reads of the memory module as the input of the decoder. However, it can not well explore the storage from the memory module since encoder branch can provide sufficient features for reconstruction. Furthermore, these methods place the memory module after the last layer of the encoder focusing more on high-level features, which are not applicable for anomaly localization requiring low-level features. Consequently, all anomaly detection methods based on AE unable to reach the pixel-level anomaly localization.

\label{sec:intro}
{
	\begin{figure*}
		\centering
		\includegraphics[width=1\linewidth]{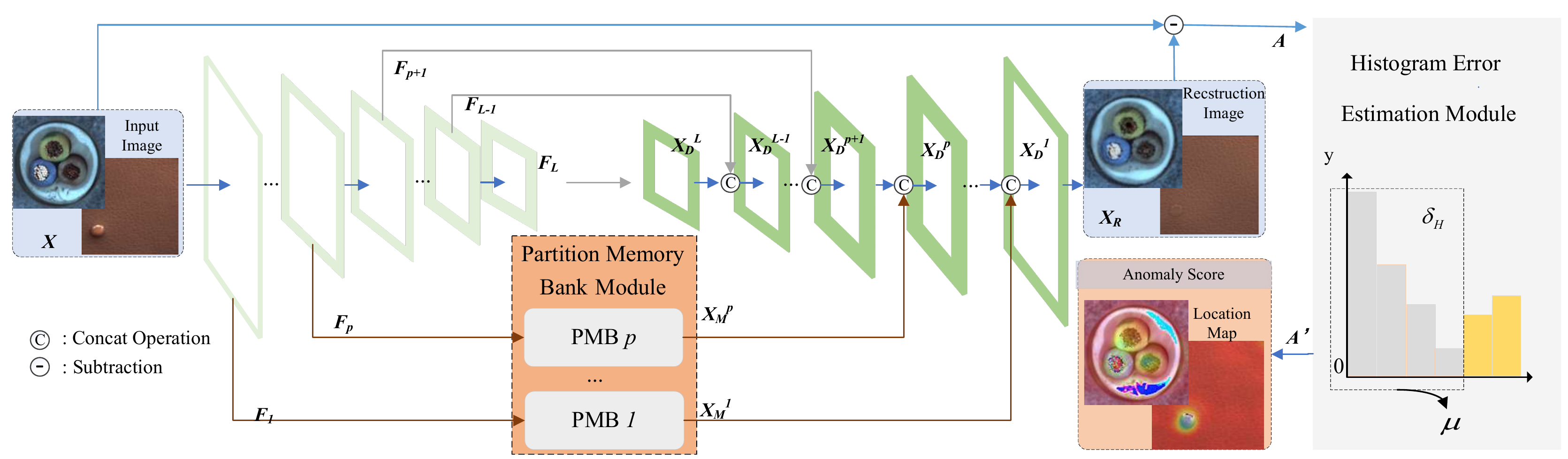}
		\caption{
			Illustration of the proposed framework. It contains the AE, the PMB module, and the histogram error estimation module. First, the $L$-layer feature map is extracted by the encoder module in the AE structure. The PMB module stores the first $P$-layer low-level features and subsequently uses skip connections. The reconstructed image $X_R$ forms an error map $A$ with the original image $X$. Second, the Histogram Error Estimation module uses $A$ to build a histogram, estimate and eliminate $\mu$, and obtain a corrected error map $A^\prime$. $A^\prime$ is used for anomaly detection and localization.}
		\label{overview}
	\end{figure*}
}
\par
Toward this end, this paper proposes a novel AE joint memory module network, as shown in Figure \ref{overview}. It utilizes a Partition Memory Bank (PMB) module to improve the abilities of learning and storing normal features, as well as a Histogram Error Estimation module to adaptively eliminate the effects of cumulative reconstruction errors caused by memory module. To preserve the critical feature information learned from the original feature map, the PMB module develops a novel query generation method that utilizes each channel feature to generate a query, as shown in Figure \ref{fig:tuhw1} (3). It allows the memory module to learn normal semantic features rather than reorganization logical features. When anomalous features as query address the PMB module, it reads the stored normal features in a semantic relationship to get normal features. Therefore, the reconstruction error becomes larger. Figure \ref{fig:fig1}(2) shows an example of a successful detection where the PMB module causes an abnormal sample B to be reconstructed as a normal sample A (the reconstruction error is large enough to be successfully detected as abnormal). Meanwhile, PMB needs to make the normal region reconstruction error as small as possible. A partition mechanism is proposed to further improve the expressiveness of the memory module, which uses multiple memory units to independently store the normal features of different partitions.
Hence, PMB module enables the small reconstruction error for normal regions while the large error for the abnormal regions, which benefits the detection of the abnormal samples. The Histogram Error Estimation module utilizes the reconstruction error maps to construct histograms. Then, it is used to adaptively estimate the reconstruction error for the normal region of each image to obtain a more effective corrected error map. More importantly, it is a non-parametric method and does not consume additional resources.

To make the PMB module put more attention on low-level features for more precise localization,
 a new scheme is developed. We directly utilize skip connections as input of decoder on high-level features and memory module reads as input on low-level features. The reads of PMB and the outputs of the skip connections are separately utilized at different layers rather than concatenated at the same layer, which allows the memory module to learn more detailed features of normal samples. Experimental results on widely-used datasets demonstrate that the proposed method obtains competitive results.
 
\par
The main contributions are summarized as follows:

(1) This paper proposes a novel Partition Memory Bank (PMB) module. The unique partition mechanism and a query generation method of PMB can effectively learn and store normal features, and achieve excellent results in anomaly detection task. 

(2) To address the challenge of more accurate anomaly localization, a new non-parametric Histogram Error Estimation module is developed to eliminate the cumulative reconstruction error, which can obtain better anomaly detection results and anomaly localization maps.

(3) The AE, PMB module, and Histogram Error Estimation module are jointly explored for their optimal compatibility. Experiments are conducted on three benchmark datasets, which shows that the proposed method can effectively solve the problem of successful reconstruction of abnormal samples.


\section{Related work}
\subsection{Unsupervised Visual Anomaly Detection}\label{A}
\par
For unsupervised anomaly detection, the detection model is trained by only normal samples, which is used to detect abnormal samples and localize abnormal regions \cite{zong2018deep,zimek2012survey}. Traditional classifiers are introduced for anomaly detection, such as SVM \cite{scholkopf2002learning} and OC-SVM \cite{OCSVM}. With the popularity of deep learning, researchers proposed deep one-class methods such as DSVDD \cite{ruff2018deep} and OCNN \cite{chalapathy2018anomaly}. The common idea of these methods is to learn the decision boundary of normal samples. For example, DSVDD learns a spherical discriminant plane to improve discrimination efficiency. There are also other methods that model the distribution features of normal samples, such as MRF  \cite{kim2009observe}, MDT \cite{mahadevan2010anomaly} and GMM \cite{xiong2011group,zimek2012survey}. However, these methods are less effective to deal with high dimensional data.

Because of the outstanding performance of transfer learning in other computer vision tasks, the pretrained network is introduced to the field of anomaly detection \cite{salehi2021multiresolution,bergmann2020uninformed,wang2021student}. In \cite{bergmann2020uninformed}, since the student network only learns latent representations of normal samples, the degree of sample abnormality is measured by the difference between latent feature representations from multiple student and teacher networks. However, they rely on additional training datasets and huge resource consumption. Some researches have proposed the employment of GAN\cite{goodfellow2014generative} to solve the anomaly detection problem. The generator in GAN is used to learn normal sample features and discriminator to identify anomalies \cite{schlegl2017unsupervised,akcay2018ganomaly,zhangnorm}. However, GAN may be unstable and tends to derive unsatisfactory results in the actual situation.

Recently, reconstruction-based methods have become popular in anomaly detection, which expects to poorly reconstruct abnormal samples and enlarges the gap of the reconstruction error of abnormal samples and normal samples \cite{hasan2016learning,zhai2016deep,zong2018deep,liu2018future,luo2017revisit,collin2021improved,ber2019,fei2020attribute}. AE  \cite{kingma2013auto} is used to reconstruct samples. For example, AE-SSIM \cite{ber2019} attempts to reconstruct normal samples directly using AE, while ARFAD \cite{fei2020attribute} learns the reconstruction features of normal samples through self-supervised tasks such as rotation. However, these methods successfully reconstruct not only normal samples but also abnormal samples, which leads to limited performance.

Different from the above methods, this work proposes a new PMB module to reconstruct successfully normal samples and poorly abnormal samples.

\begin{figure*}
	\centering
	\includegraphics[height=0.35\linewidth,width=1\linewidth]{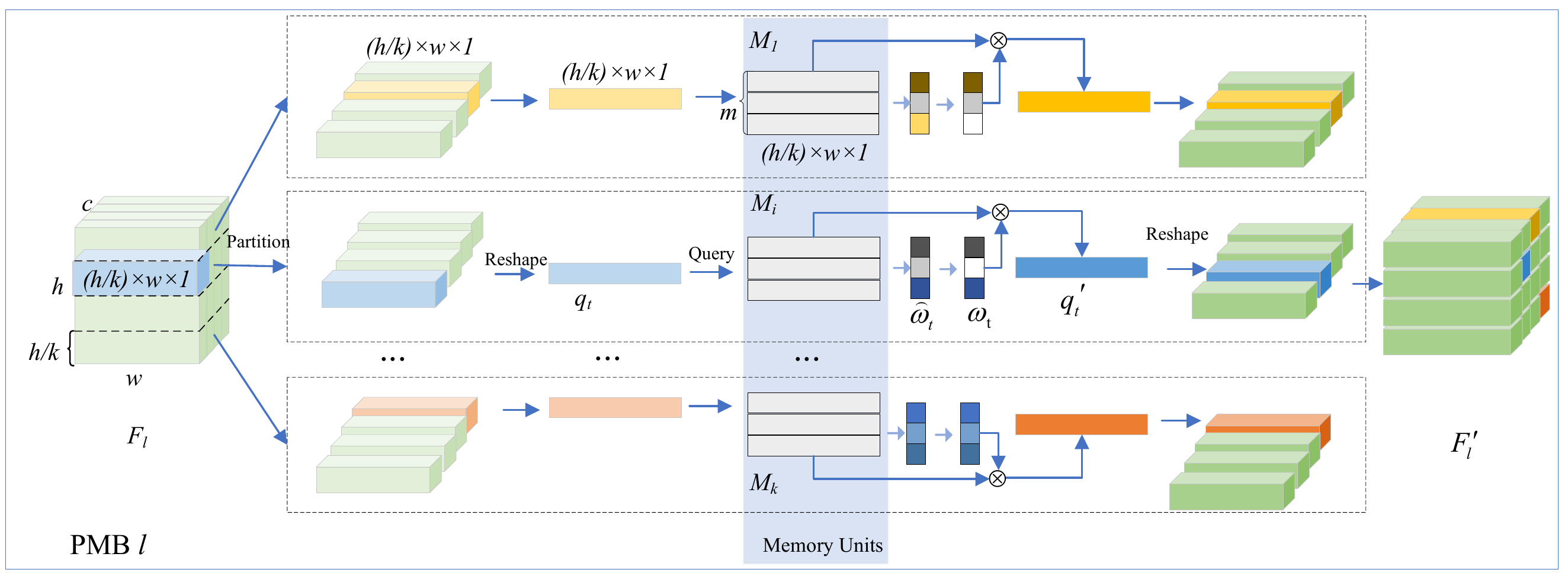}
	\caption{The framework of the proposed PMB module. For each layer of feature map $F_l$, the partition mechanism causes the feature map to be divided into separate partitions. Each partition has a separate memory unit to store normal features. Each query addresses the memory unit in an attentional way to obtain the valid normal feature reads as input to the decoder.}
	\label{memory}
	
\end{figure*}
\subsection{Memory Network}
\par
The memory network was initially adapted to the field of natural language processing \cite{weston2014memory,hochreiter1997long,kumar2016ask,fan2019heterogeneous,DBLP:journals/tcsv/WenHL21}. Recently, Some existing works apply memory networks to anomaly detection \cite{park2020learning,wang2021cognitive}.  Gong et al. \cite{gong2019memorizing} proposes MemAE, which uses a memory module to reduce the reconstruction ability of AE for anomalous samples. However, due to its simple memory module and the shortcomings of the combination method, it is difficult to be effective for tiny anomaly sample detection and localization tasks. Park et al. \cite{park2020learning} proposes a combination network of AE and memory module for video anomaly detection. It constructs an update method for the memory items of the memory module and concatenates the outputs of AE and memory module as the input of the decoder. However, it is not suitable for the localization challenges of complex scenes because it focuses more on high-level features. Second, the output of AE as the input of the decoder results in the limitation of features stored in the memory module.
	Wang et al. \cite{wang2021cognitive} introduces an additional evaluation network on the MemAE framework as a discriminator to detect whether a sample is anomalous or not. However, it is difficult to achieve anomalous region localization with discriminators.

 The most related study to the proposed method is DAAD. \cite{hou2021divide}. It uses the method in Figure \ref{fig:tuhw1} (2) to generate memory query of size $a\times b \times c$ ( $a \le h$, $b \le w$ ) and uses a discriminator to identify whether the samples are anomalous or not. DAAD uses larger query (generally $8\times 8 \times c$) to make anomalies and normals that do not share the same block pattern. However, it destroys the integrity of the features learned by the convolutional network, resulting in the memory module learning reorganized logical information. Therefore, abnormal features can also be logically fitted to anomalies such that reconstruction errors are small. The proposed PMB module employs novel query of size $h\times w\times 1$ (Figure \ref{fig:tuhw1}(3)) which enables the memory module to store normal feature learned from the convolutional network. Therefore, all queries can address the PMB module to get normal features. Unlike DAAD, it is not degrade the reconstruction capability of AE, but make the memory module unable to address abnormal features. In addition, DAAD introduces discriminator with adversarial loss to solve the problem of incorrect discrimination caused by cumulative errors, which results in additional resource consumption and the failure of anomaly localization. In this paper, we not only ensure that the normal features are well reconstructed, but also the anomaly localization results can be derived from the error map without additional resources.

\section{The Proposed Method}

\subsection{Overview}

\par
This work proposes to solve the unsupervised anomaly detection problem based on the reconstruction scheme. The overview of the proposed framework is shown in Figure \ref{overview}. The proposed framework contains three basic structures: 1) an AE, 2) a PMB module, and 3)  a Histogram Error Estimation module. 
\par
Given the input sample $X$, the encoder first extracts the multi-scale latent representations $F=\{F_1,F_2,\ldots,F_L\} $. To make full utilization of the representation information provided by the memory modules and allow the memory module to focus on low-level features, we utilize the output of the PMBs at layer $1$ to layer $p$ of the encoder and the output of the subsequent $p+1$ to $L$ layers with the skip connection. Therefore, $p$ PMBs are designed to learn and store feature maps of different scales independently. The reads of the proposed PMB module are represented by $X_M = \{X_M^1,X_M^2,\ldots,X_M^p\}$. Then, the channel concatenation between $X_M$  and output of previous decoder $X_D$  as the input of the next decoder.  The reconstructed image $X_R$ is obtained after the decoder. The original difference image $A$ is obtained by using Eq. \ref{equa_4:A},
\begin{equation}
	A = |{{\rm{X}} - {X_R}}| \label{equa_4:A}.
\end{equation}
Besides, the Histogram Error Estimation module is developed to eliminate reconstruction errors in normal regions to obtain a more accurate corrected error map $A^\prime$ for anomaly localization and detection. The anomaly scores and anomaly localization maps of the samples are obtained based on $A^\prime$.

\subsection{Partition Memory Bank Module}
\par

\par 
As is shown in Figure \ref{memory}, The PMB module is proposed to learn and store the potential feature representations of normal samples. It achieves the challenge of anomalous sample reconstruction error becoming large by exploring the proposed joint partition mechanism and a new query generation method.

To generate the query with critical feature information, a novel query generation method is developed, as shown in Figure \ref{fig:tuhw1} (3).
It generates queries directly from each channel of the feature map. Thus, each query contains semantic information for a single channel, allowing the PMB module to store only normal semantic information.
In addition, to enhance the learning ability and expressiveness of the memory module and to make the reconstruction error of normal regions small, a partition mechanism is proposed. It introduces multiple local units to store the normal features of different regions separately. Each memory unit is capable of storing detailed feature information for individual regions. Multiple memory units make the expressiveness of the memory module enhanced.

Specifically, for each feature map $F_l$ with the size of $h\times w\times c$ from $F$, it is partitioned by using the proposed query generation approach shown in Figure \ref{memory}. Along the `$h$' dimension of the feature map, $F_l$ is partitioned into $k$ regions with the size of $\frac{h}{k}\times w\times c$ , each of which is stored and queried by a separate memory unit. The feature map of individual channel in each partition is flattened to one $\frac{h}{k}\times w$ dimensional vector. Thus, we can get query sets $Q=\{\boldsymbol{q_t} | t\in[1,c]\}$  in each partition. Each PMB contains $k$ memory units $\boldsymbol{M}=\{\boldsymbol{M_1},\boldsymbol{M_2},\ldots,\boldsymbol{M_k}\}$, each memory unit contains $m$ memory items, and the size of the memory items is the same as the query $q_t$. Finally, as shown in Fig. \ref{memory} (4), suppose $\boldsymbol{q_t}$ is a query for $i$-th partition of $F_l$, $\boldsymbol{q_t}$ and the corresponding memory items in $\boldsymbol{M_i}$ are first normalized to improve the accuracy of the attention weight. The normalization operations are as follows:

\begin{equation}
	\boldsymbol{\widehat{q_t}}   = norm\left( {{\boldsymbol{q_t} }} \right)
	,
	\boldsymbol{\widehat {M_i^j}}   = norm\left( \boldsymbol{{M_i^j}}  \right), j \in \left[ {1,m} \right],
\end{equation}
where $norm(\cdot)$ represents the $l_2$-norm. $	\boldsymbol{\widehat{q_t}}$ represents the normalized the query $\boldsymbol{q_t}$. $\boldsymbol{\widehat {M_i}}$ represents the memory items after normalization.  Then the similarity between $\boldsymbol{\widehat{q_t}}$ and the memory item $\boldsymbol{\widehat {M_i}}$ is used to calculate the attention weight $\boldsymbol{\widehat {\omega _t}}$ as follows:
\begin{equation}
	{\widehat{ \omega_t^j}}=\frac{exp(<\boldsymbol{\widehat{q_t}},\boldsymbol{\widehat{ M_i^j}}>)}{\sum\limits_{u = 1}^m{exp(<\boldsymbol{\widehat{q_t}},\boldsymbol{\widehat{M_u^j}}>)}}
\end{equation}
Here $<\cdot,\cdot>$ denotes the cosine similarity. To avoid too small weight of the query, this paper proposes to filter the weights by introducing a threshold, which can alleviate the reconstruction with most of memory items \cite{gong2019memorizing}. Consequently, when abnormal features are used as queries, the reads are difficult to fit and the abnormal regions are then difficult to be reconstructed.
\begin{equation}
	\omega_t^j=max\left(0,\widehat{ \omega_t^j}-\delta_m\right),
\end{equation}
where ${{\delta _m}}$ represents the threshold.
By using the memory module, we can obtain a new feature vector $\boldsymbol{q_t^\prime}$ for each $\boldsymbol{q_t}$ by using the output of the memory unit $\boldsymbol{M_i}$.
\begin{equation}
	\boldsymbol{q_t^\prime}=\boldsymbol{\omega_t}\ \cdot\ \boldsymbol{M_i}  
\end{equation}
Then a new feature map $F^\prime$ can be obtained by re-arranging all $\boldsymbol{q_t^\prime}$  and fed into the decoder to generate the reconstructed image $X_R$. It is worth noting that the stability of the reconstructed network can be improved by using $\boldsymbol{q_t^\prime}$ since the value of $\boldsymbol{q_t^\prime}$ does not change much.

\subsection{Histogram Error Estimation Module}

\par
For each image, the original error map $A$ is obtained by using Eq. \ref{equa_4:A}, in which the value of each pixel represents the reconstruction error. In the proposed approach, the reconstruction error comes from the proposed PMB module and AE. The PMB module learns and stores the normal features, which enables to guarantee that the reconstruction errors of normal regions are obviously smaller than ones of abnormal regions. However, due to the presence of the memory module, AE is difficult to achieve zero-error reconstruction in the normal region. 
These accumulative reconstruction errors can result in false detection from the original error map. Therefore, we need to further mitigate the impact of accumulative reconstruction errors on detection and localization performance.

In this paper, we explore a simple, yet effective Histogram Error Estimation module to estimate the error $\mu$. As shown in Figure \ref{overview}, the  histogram of error map is first constructed. Due to the small reconstruction error of normal region, the pixels in normal region always lie on the left side of histogram. Therefore, we can quickly find normal pixels in the histogram. Then, a percentage $\delta_H$ of small reconstruction errors are chosen to estimate $\mu$ as shown in the dashed box of the histogram in Figure \ref{overview}. The average value of errors of these selected pixels is utilized to estimate $\mu$. Then, the corrected difference image  $A^\prime$ is obtained as follows:

\begin{equation}
	{{A}^\prime }{\rm{ = }}\left\{ {\begin{array}{*{20}{c}}
			{A- \mu ;}&{A > \mu }\\
			{0;}&{A \le \mu } 
	\end{array}} \right.\label{A'}
\end{equation}

The anomaly score and localization map of the samples are derived from $A^\prime$. $A^\prime$ is adopted as the localization map, while the anomaly score is calculated by the sum of the error values of the corrected difference image $A^\prime$.

\begin{table*}[]
	\centering
	\caption{
		Comparison of the proposed method and anomaly detection methods on MVTec AD dataset with AUROC (\%). The best results are marked in bold. `\checkmark' means that the method additionally introduces large datasets such as IMAGENET.}
	\renewcommand\arraystretch{1.5}
	\setlength{\tabcolsep}{1mm}{
		\begin{tabular}{@{}c|c|c|ccccccccccccccc|c@{}}
			\hline\scriptsize &
			\scriptsize Method&\begin{tabular}[c]{@{}c@{}} \scriptsize Extra\\ \scriptsize Datasets \end{tabular}&\scriptsize Bottle&\scriptsize Hazelnut&\scriptsize Capsule& \begin{tabular}[c]{@{}c@{}} \scriptsize Metal\\ \scriptsize Nut\end{tabular}&\scriptsize Leather&\scriptsize Pill&\scriptsize Wood&\scriptsize Carpet&\scriptsize Tile&\scriptsize Grid&\scriptsize Cable&\scriptsize Transistor&\scriptsize Toothbrush&\scriptsize Screw&\scriptsize Zipper&\scriptsize Mean\\
			\hline
			&CNN-Dict\cite{napoletano2018anomaly}&\checkmark&78&72&84&82&87&68&91&72&93&59&79&66&77&87&76&78\\
			
			&MKDAD\cite{salehi2021multiresolution}&\checkmark&99.4&98.4&80.5&73.6&95.1&82.7&94.3&79.3&91.6&78.0&89.2&85.6&92.2&83.3&93.2&87.7\\
			
			Pre-trained&SPADE \cite{cohenspade}&\checkmark		&-&-&-&-&-&-&-&-&-&-&-&-&-&-&-&85.5\\
			&FAVAE \cite{davidAL}&\checkmark& 99.9&99.3&80.4&85.2&67.5&82.1&94.8&67.1&80.5&97.0&95.0&93.2&95.8&83.7&97.2&87.9\\
			&Cutpaste\cite{li2021cutpaste}&\checkmark&100&99.7&94.3&98.7&100&91.3&99.8&100&98.9&98.8&93.9&95.6&92.8&86.0&99.9&\textbf{96.6}\\
			\hline
			&CAVGA-$D_w$\cite{venkataramanan2020attention}&\checkmark&93&90&89&81&80&93&89&80&81&79&86&80&96&79&95&86\\
			&Puzzle-AE\cite{salehicorr}&-&94:2&91:2&66:9&66:3&72:9&71:6& 89:5&65:7&65:5&75:3&87:9&85:9&97:8&57:8&75:7&77.6\\
		AE	&ARFAD\cite{fei2020attribute}&-&94.1&85.5&68.1&66.7&86.2&78.6&92.3&70.6&73.5&88.3&83.2&84.3&100.0&100.0&87.6&83.9\\
			&MemAE\cite{gong2019memorizing}&-&95.4&89.1&83.1&53.7&61.1&88.3&95.4&45.4&63.0&94.6&69.4&79.3&97.2&99.2&87.1&80.2\\
			&AESc\cite{icprCollinV20}&-&98.0&94.0&74.0&73&89.0&84.0&95.0&89.0&99.0&97.0&89.0&91.0&100&74.0&94.0&89.0\\
			&\textbf{OURS}&-&95.2&99.4&82.3&84.5&94.5&86.1&100.0&93.1&97.2&97.1&85.6&92.4&95.8&97.0&77.3&\textbf{91.8}\\
			\hline

	\end{tabular}}
	\vspace{0.2em}
	\label{tab:mvtec}
\end{table*}

\begin{table*}[h]
	\setlength{\abovecaptionskip}{0cm}
	\centering
	\caption{AUROC of the proposed method and the compared methods for anomaly detection on MNIST.}
	\renewcommand\arraystretch{1.6}
	\setlength{\tabcolsep}{1.6mm}{
		\begin{tabular}{c|ccccccccc}\hline
			Method&DSVDD\cite{ruff2018deep}&	CapsNetPP\cite{li2020exploring}&OCGAN\cite{perera2019ocgan}&LSA\cite{abati2019latent}&MemAE\cite{gong2019memorizing}&		OCSVM\cite{OCSVM}&ARAE\cite{salehi2020arae}&\textbf{OURS}& MKDAD \cite{salehi2021multiresolution} \\ 
			\hline
			Pre-trained model&-&-&-&-&-&-&-&-&\checkmark\\ \hline
			AUROC&94.8&97.7&97.5&97.5&97.5&96.0&97.5&98.1& \textbf{99.35} \\
			\hline
	\end{tabular}}
	
	\label{tab:mnist}
	\vspace{0.1em}
\end{table*}

\subsection{Loss Function}
\par
To train the proposed model, the quality of the reconstructed image and the intuitive feeling of the human visual system are jointly explored.
The reconstruction loss $\mathcal{L}_{MSE}$ and  structural similarity index measure (SSIM) \cite{2004WangBSS04} loss ${\mathcal L}_{{\rm{SSIM}}}$ are
introducesd , which are defined as:
\begin{equation}
	\mathcal {L}_{MSE} = \left \| {X - {X_R}} \right \| _2 
\end{equation}

\begin{equation}
	{{\mathcal L}_{{\rm{SSIM}}}} {\rm{ = }}\frac{1}{{h \times w}}\sum\limits_{i = 1}^h {\sum\limits_{j = 1}^w {(1 - {\rm{SSIM}}_s^{(i,j)}(X,{X_{\rm{R}}}))} }  
\end{equation}
Here $h$ and $w$ represent the height and width of the image. ${\rm{SSIM}}_s^{(i,j)}(X,{X_{\rm{R}}})$ represents structural similarity index between the image $X$ and image $X_R$ at the center $(i,j)$ position with kernel size of s.

This paper proposes to jointly explore the above loss to obtain the total training loss as follows:
\begin{equation}
	{\mathcal L} {\rm{ = }} {{\mathcal L}_{MSE}} + {\lambda _{{\rm{SSIM}}}} \cdot {{\mathcal L}_{SSIM}},
\end{equation}
where ${\lambda _{SSIM}}$ is the hyperparameter of the model that measures the importance of the SSIM loss.

\section{Experiments}
This section conducts extensive experiments to verify the effectiveness of the proposed method with the comparison to several state-of-the-art methods.

\subsection{Experimental Setting}

\par
\noindent\textbf{Datasets.} To validate the effectiveness of the proposed method, experiments are conducted on three widely-used datasets. 

(1) \emph{\textbf{MVTec AD}} \cite{bergmann2019mvtec}. The MVTec AD dataset contains 5354 high-resolution color images in 15 different categories.  Among them, There are 5 categories for textured images, such as ``wood'' or ``leather''. The other 10 categories contain non-textured objects, such as ``cable''. The challenge of this dataset comes from the fact that normal images and abnormal images come from the same category.  The difference between the abnormal sample and the normal sample is subtle, such as scratches on the ``leather''. Each category contains a dataset including only normal images for training and a dataset including normal and various abnormal images for testing. 

(2) \emph{\textbf{MNIST}} \cite{lecun1998mnist}. The dataset contains 10 different types of low-resolution grayscale images.  A total of 70,000 images.  60,000 pictures for training, and 10,000 pictures for testing. In anomaly detection, a category is used as a normal sample and the remaining 9 categories are used as anomaly samples (called out-of-distribution detection \cite{yangOOD}).

(3) \emph{\textbf{Retinal-OCT}} \cite{gholami2020octid}
 It is a medical dataset, which contains normal (healthy) retinal CT images and 3 categories of abnormal (damaged) retinal CT images.

\par
\noindent
\textbf{Implementation Details.} In experiments, each image is resized into the size of  $128\times128$. The UNet \cite{ronneberger2015u}-like network is utilized as the autoencoder in this paper. The learning rate is set to $0.002$. For MVTec AD, $L$ is set to $5$ and $p$ is set to $4$. The number of $k$ in each PMB is set to $( 32, 16, 16, 8 )$ for five texture categories and $\left( 16, 16, 16, 8 \right)$ for other object  categories. Accordingly, the number of memory items $m$ is $( 150, 200, 200, 200)$. For MNIST, $L$ and $p$ are both set to $3$. The number of partitions $k$ is set to $( 1, 1, 1 )$  and $m$ is set to $( 10, 10, 10 )$, respectively. For Retinal-OCT, we set $L$ and $p$ to $4$ and $3$, respectively. $k$ in each PMB is set to $( 2, 2, 2 )$, respectively. The number of memory items $m$ is set to $( 10, 10, 10 )$, respectively. The batch size is set to $20$ for the MVTec AD and Retinal-OCT datasets, as well as $64$ for MNIST. For the hyper-parameters in the proposed formulation, follow \cite{wang2021cognitive}, $\delta_m$ is set to $\frac{1}{m}$, $\delta_H$ and $\lambda_{SSIM}$ are set to $50\%$ and $0.1$, respectively.

\par
\noindent
\textbf{Evaluation.}
Following \cite{park2020learning,akcay2018ganomaly},  the area under the curve (AUC) of the receiver operating characteristic (ROC) of the image level and pixel level are used to measure the performance of model.

\begin{table*}
	\centering
	\setlength{\abovecaptionskip}{0cm}
	\caption{AUROC of the proposed method and the compared methods for anomaly detection on Retinal-OCT.}
	\renewcommand\arraystretch{1.5}
	\setlength{\tabcolsep}{1.6mm}{
		\begin{tabular}{c|ccccccc c}\hline
			Method&DSVDD\cite{ruff2018deep}&AnoGan\cite{schlegl2017unsupervised}& 
			VAE-GAN\cite{baur2018deep}&Cycle-GAN\cite{zhu2017unpaired}&	GANomaly\cite{akcay2018ganomaly}&P-Net\cite{zhou2020encoding}&MKDAD\cite{salehi2021multiresolution}&\textbf{OURS}\\
			\hline
			pre-trained model&-&-&-&-&-&-&\checkmark&-\\ \hline
			AUROC&74.4&84.81&90.64&87.39&91.96&92.88&97.01&\textbf{98.00} \\
			\hline
	\end{tabular}}
	
	\label{tab:Retina}
	\vspace{0.2em}
\end{table*}
\begin{table*}[htb]
	\centering
	\setlength{\abovecaptionskip}{0cm}
	\caption{AUROC of the proposed method and the compared methods for anomaly localization on MVTec AD. }
	\renewcommand\arraystretch{1.5}
	\setlength{\tabcolsep}{1mm}{
		\begin{tabular}{c|c|ccccccccccccccc|c}\hline
			\scriptsize Method&\begin{tabular}[c]{@{}c@{}} \scriptsize Extra\\ \scriptsize Datasets \end{tabular}&\scriptsize Bottle&\scriptsize Hazelnut&\scriptsize Capsule&\scriptsize MetalNut&\scriptsize Leather&\scriptsize Pill&\scriptsize Wood&\scriptsize Carpet&\scriptsize Tile&\scriptsize Grid&\scriptsize Cable&\scriptsize Transistor&\scriptsize Toothbrush&\scriptsize Screw&\scriptsize Zipper&\scriptsize Mean\\
			\hline
			AE-SSIM\cite{ber2019}&-&93&97&94&89&78&91&73&87&59&94&82&90&92&96&88&87\\
			AE-$L_2$\cite{ber2019}&-&86&95&88&86&75&85&73&59&51&90&86&86&93&96&77&82\\
			AnoGAN\cite{schlegl2017unsupervised}&-&86&87&84&76&64&87&62&54&50&58&78&80&90&80&78&74\\
			CNN-Dict\cite{napoletano2018anomaly}&\checkmark&78&72&84&82&87&68&91&72&93&59&79&66&77&87&76&78\\
			MKDAD\cite{salehi2021multiresolution}&\checkmark&96.3&94.6&95.8&86.4&98.1&89.6&84.8&95.6&82.8&91.8&82.4&76.5&96.1&95.9&93.9&90.7\\
			CAVGA-$D_u$\cite{venkataramanan2020attention}&-&-&-&-&-&-&-&-&-&-&-&-&-&-&-&-&85.0\\
			UTAD\cite{liu2021unsupervised}&-&-&-&-&-&-&-&-&-&-&-&-&-&-&-&-&90.0\\
			CAVGA-$D_w$\cite{venkataramanan2020attention}&\checkmark&-&-&-&-&-&-&-&-&-&-&-&-&-&-&-&92.0\\
			
			\hline
			\textbf{OURS}&-&93.7&92.5&92.1&85.8&96.7&91.1&86.5&92.3&90.7&94.3&94.2&80.8&97.5&97.7&95.4&\textbf{92.1}\\
			\hline
	\end{tabular}}
	
	\label{tab:MVTEC segmentation}
\end{table*}

\subsection{Experimental Results}

\begin{figure}
	\centering
	\includegraphics[width=1\linewidth]{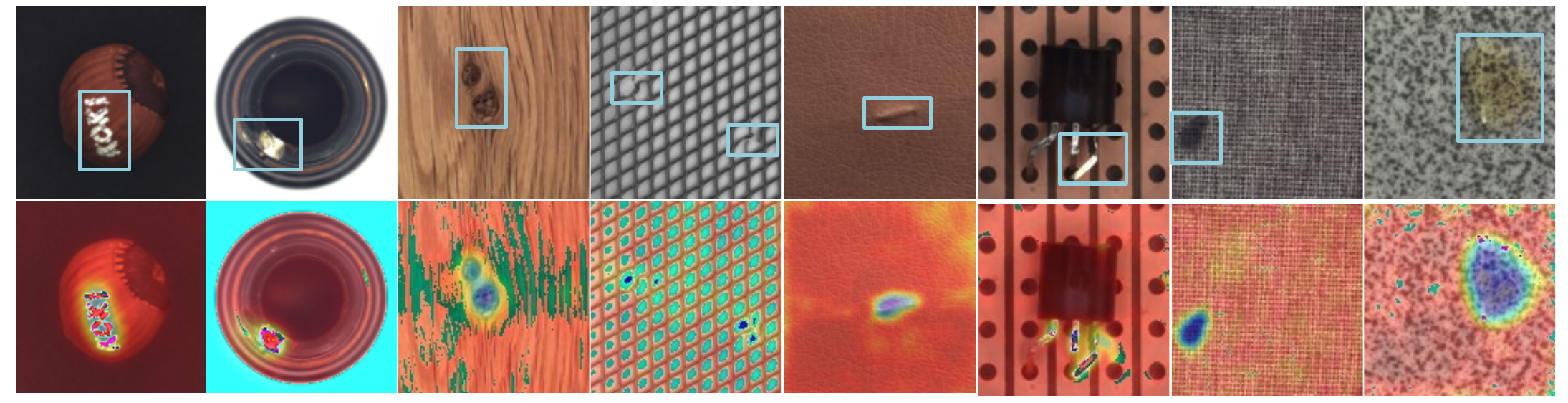}
	\caption{Illustration of the localization heatmap. Anomaly images are shown in the top row and the corresponding weighted heatmaps are presented in the bottom row. The locations
		of abnormal regions are denoted by the rectangular
		box.}
	\label{fig:localizatiom-map}
\end{figure}

\par
\noindent
\textbf{Anomaly detection results.} The anomaly detection results on these three datasets and the comparison with the state-of-the-art methods are shown in Table \ref{tab:mvtec}, \ref{tab:mnist} and \ref{tab:Retina}, respectively.

\vspace{0.5em}

(1) \textbf{MVTec AD.}
For the MVTec AD dataset, this paper compares the state-of-the-art methods with distribution-based methods and AE methods. The distribution-based approach used pre-trained models with ImageNet dataset \cite{feifeiimagenet}, such as CNN-Dict \cite{napoletano2018anomaly}, MKDAD \cite{salehi2021multiresolution}, SPADE \cite{cohenspade}, FAVAE \cite{davidAL}, Cutpaste \cite{li2021cutpaste}.
Among them, MKDAD uses the VGG-16 \cite{simonyan2014very} model pre-trained on ImageNet, SPADE uses the ResNet-18 \cite{he2016deep} model pre-trained on ImageNet, and CutPaste uses the EfficientNet (B4) \cite{tan2019efficientnet} model pre-trained on ImageNet. AE methods include CAVGA-D$_w$ \cite{venkataramanan2020attention}, Puzzle-AE \cite{salehicorr}, ARFAD \cite{fei2020attribute}, memAE \cite{gong2019memorizing}, AESc \cite{icprCollinV20}. For example, Puzzle-AE introduces the complex self-supervised task of puzzle reduction, ARFAD introduces the self-supervised task of rotation and coloration, and MemAE introduces the traditional memory module. 

From Table \ref{tab:mvtec}, the proposed method achieves mean AUROC 91.8\%. It is lower than Cutpaste due to the fact that Cutpaste uses a better pre-trained model and uses forged anomalous samples for training. However, the proposed method outperforms the remaining methods using pre-trained models by 4\%-10\%, validating the effectiveness of the AE combined with the PMB module. Among all AE methods, our method achieves state-of-the-art performance. It outperforms the CAVAG-D$_w$ method that introduces real anomalous samples for training by 6\%, validating that our method also performs very well without relying on anomalous samples. It is 13\% and 8\% higher than Puzzle-AE and ARFAD methods, indicating that our method can achieve excellent performance without designing complex image preprocessing operations. It outperforms the traditional memory model MemAE method by 11\%, validating that the proposed PMB module stores normal features more efficiently.

\vspace{0.5em}

(2) \textbf{MNIST. }  
For the MNIST dataset, this paper compares the state-of-the-art methods including DSVDD\cite{ruff2018deep}, CapsNetPP\cite{li2020exploring}, OCGAN\cite{perera2019ocgan}, LSA\cite{abati2019latent}, MemAE\cite{gong2019memorizing}, OCSVM\cite{OCSVM}, ARAE\cite{salehi2020arae}, , and MKDAD \cite{salehi2021multiresolution}. From Table \ref{tab:mnist}, it can be found that the proposed method achieves mean AUROC 98.1\%\cite{yangOOD} in the out-of-distribution detection setting. The advanced performance shows the superiority of the proposed method. It is slightly lower than the pre-trained method MKDAD method but far outperforms the remain distribution methods. Besides, compared with the traditional memory module MemAE, the proposed method outperforms MemAE, which validates the effectiveness of the novel query method and partition mechanism.

\vspace{0.5em}

 (3) \textbf{Retinal-OCT.} In the medical anomaly detection task, the proposed method achieves significant performance advantages. We compare with DSVDD \cite{ruff2018deep}, AnoGan \cite{schlegl2017unsupervised},  
 VAE-GAN \cite{baur2018deep}, Cycle-GAN\cite{zhu2017unpaired}, 	GANomaly \cite{akcay2018ganomaly}, P-Net \cite{zhou2020encoding}, and MKDAD \cite{salehi2021multiresolution} on Retinal-OCT dataset. The results are shown in Table \ref{tab:Retina}. It can be found that the proposed method achieves the state-of-the-art performance with AUROC of 98.0\%. It not only outperforms recent methods using GAN networks \cite{baur2018deep}, \cite{zhu2017unpaired}, but even outperforms the MKDAD method. Our method does not rely on additional training data inductive bias specific to the pre-trained model. It is more suitable for medical image anomaly detection.
\vspace{1.5em}

\begin{figure}
	\centering
	\includegraphics[width=1\linewidth]{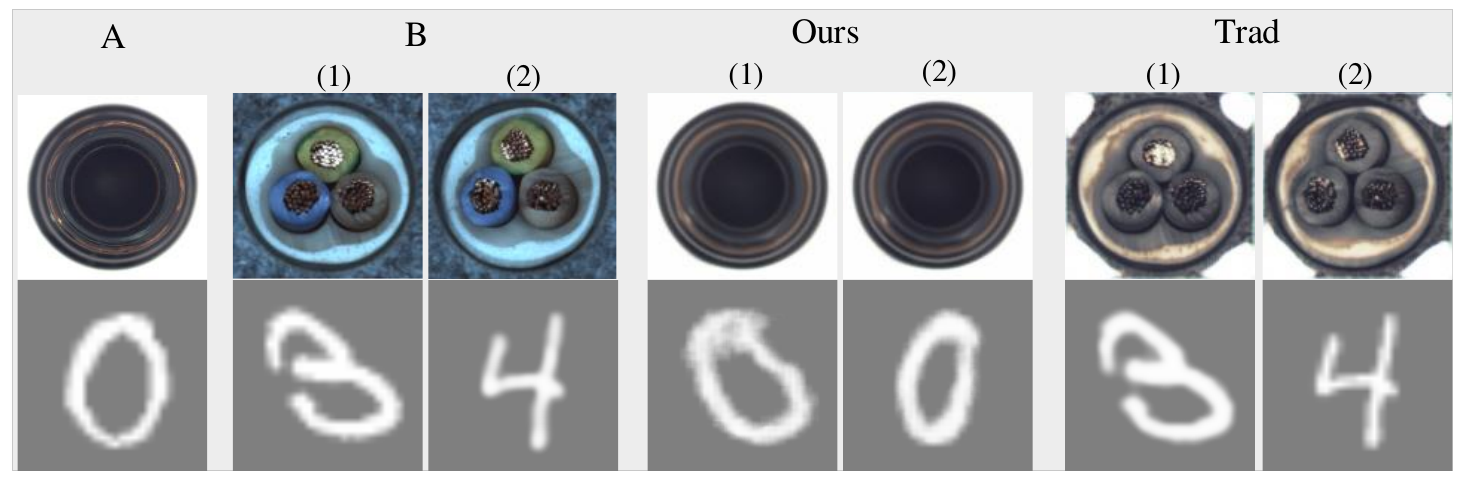}
	\caption{Illustrations of reconstructed images by using the proposed PMB module or the traditional memory module. 
		`A' represents normal samples and `B' represents abnormal samples, `Ours' shows the test results of the PMB module and `Trad' shows the test results of the traditional memory module.
		 The reconstruction error of the proposed PMB module (`Ours') reconstructing anomalous sample `B' is very large and can be effectively detected. However, the traditional module has a small reconstruction error for abnormal samples(`Trad').}
	\label{fig:inter}
\end{figure}

\par
\noindent
\textbf{Anomaly localization results.}  As far as we know, none of the recent AE (including memory module) anomaly detection methods achieve pixel-level anomaly detection (anomaly localization task), which may be due to the fact that these methods weak AE reconstruction capability resulting in cannot guarantee small reconstruction error for normal region. However, the proposed method does not weaken the reconstruction capability of the model but makes the anomalous features unaddressable from the PMB module thereby increasing the reconstruction error of the anomalous samples. Moreover, the unfavorable error is eliminated by using the Histogram Error Estimation module to further enlarge the difference. The results of anomaly localization on the MVTec AD dataset are shown in Table \ref{tab:MVTEC segmentation}. As a result, our method achieves 92.1\% performance in anomaly localization.

 Compared with AE-based methods such as AE-SSIM \cite{ber2019} and AE-$L_2$ \cite{ber2019}, the proposed method gains 5\%-10\% improvement. Because the proposed PMB module can store the learned normal sample features, effectively making the anomalous region reconstruction error larger. Compared with the two-stage method UTAD \cite{liu2021unsupervised}, the proposed method is simpler and more effective in training, and better localization results are obtained by exploring the corrected differential image $A^\prime$. Compared with the weakly supervised method CAVGA-$D_w$ \cite{venkataramanan2020attention} which uses additional abnormal region annotation information, the proposed method achieves better results. It shows that the proposed method outperforms the method with weakly supervised settings in the unsupervised setting, fully validating its superiority. To further demonstrate the localization performance of the proposed method, some examples of anomaly localization are illustrated in Figure \ref{fig:localizatiom-map}. It can be observed that the proposed method can well solve the problem of abnormal region localization.

\begin{figure*}
	\centering
	\includegraphics[width=1\linewidth]{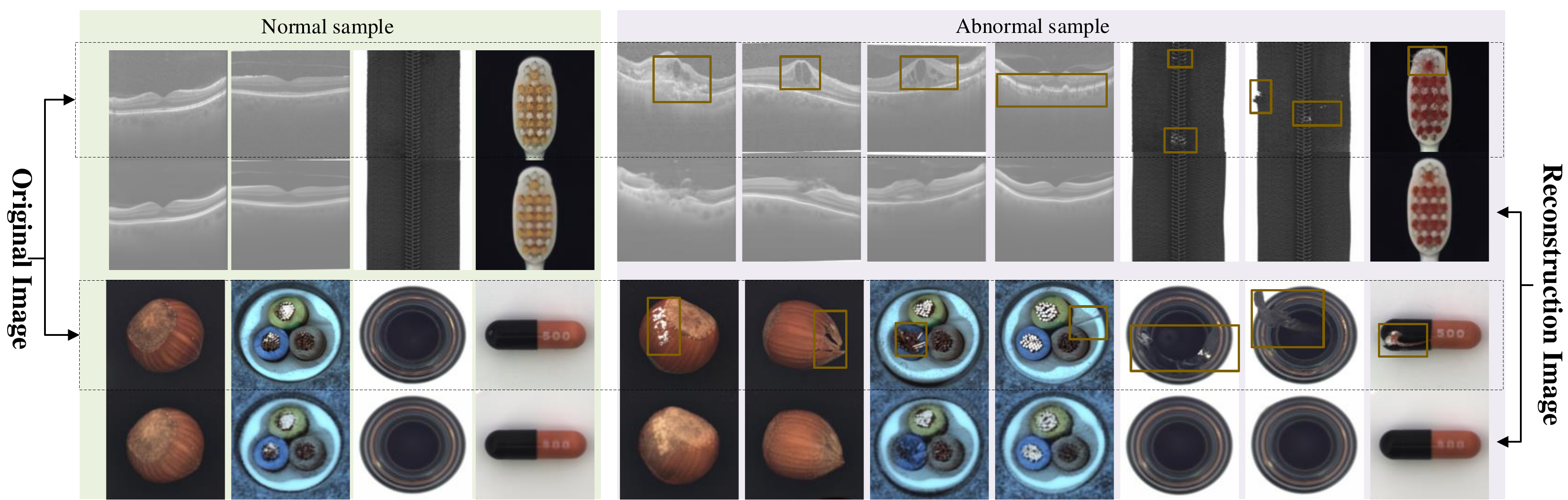}
	\caption{Some examples of reconstruction of abnormal and normal samples. The original images and the reconstructed images are listed in the top row and the bottom row, respectively. }
	\label{fig:moreexample}
\end{figure*}

\begin{table*}[]
	\centering
	\caption{
		The results of the comparison between the proposed method and Baselines. The best results are marked in bold. The results of `AE+skip' are from \cite{hou2021divide}. `W/O H' represents the proposed method without Histogram Error Estimation module.}
	\renewcommand\arraystretch{1.5}
	\setlength{\tabcolsep}{1.6mm}{
		\begin{tabular}{@{}c|ccccccccccccccc|c@{}}
			\hline
			\scriptsize Method&\scriptsize Bottle&\scriptsize Hazelnut&\scriptsize Capsule& \begin{tabular}[c]{@{}c@{}} \scriptsize Metal\\ \scriptsize Nut\end{tabular}&\scriptsize Leather&\scriptsize Pill&\scriptsize Wood&\scriptsize Carpet&\scriptsize Tile&\scriptsize Grid&\scriptsize Cable&\scriptsize Transistor&\scriptsize Toothbrush&\scriptsize Screw&\scriptsize Zipper&\scriptsize Mean\\
			\hline

			AE-SSIM\cite{ber2019}&88&54&61&54&46&60&83&67&52&69&61&52&74&51&80&63\\
			AE-$L_2$\cite{ber2019}&80&88&62&73&44&62&74&50&77&78&56&71&98&69&80&71\\
			AE+skip &71.3&82.8&74.7&33.6&57.0&85.3&97.7&38.5&98.6&87.9&57.9&74.9&74.2&100.0&69.6&73.6\\

			MemAE\cite{gong2019memorizing}&95.4&89.1&83.1&53.7&61.1&88.3&95.4&45.4&63.0&94.6&69.4&79.3&97.2&99.2&87.1&80.2\\
			DAAD\cite{hou2021divide}&97.5&89.3&86.6&55.2&62.8&89.8&95.7&67.1&82.5&97.5&72.0&81.4&98.9&100.0&90.6&84.5\\
			DAAD+\cite{hou2021divide}&97.6&92.1&76.7&75.8&86.2&90.0&98.2&86.6&88.2&95.7&84.4&87.6&99.2&98.7&85.9&89.5\\
			\hline
			\textbf{OURS (W/O H)}&98.7&94.8&72.5&75.3&97.2&69.3&100.0&95.0&88.1&99.1&81.6&90.8&87.5&100&82.43&88.8\\
			\textbf{OURS}&95.2&99.4&82.3&84.5&94.5&86.1&100.0&93.1&97.2&97.1&85.6&92.4&95.8&97.0&77.3&\textbf{91.8}\\
			\hline
	\end{tabular}}
	\vspace{0.2em}
	\label{tab:AEcom}
\end{table*}

\subsection{Ablation study}
\par
\noindent
\textbf{The impact of the different memory models. } 
To validate the advantages of the proposed PMB module over the traditional memory module for anomaly detection, we used the OOD detection setting to demonstrate the reconstruction content: one category as the normal samples and the remaining categories as abnormal samples. We conducted visualization studies in two datasets (MVTec AD and MNIST) separately. The results are illustrated in Figure \ref{fig:inter}. For example, by using category `0' as the normal samples for training, the samples `3' and `4' were reconstructed during the test. It can be observed that since the PMB module cannot successfully reconstruct the samples of class `3' and `4', the reconstruction results are shown as class `0'. The abnormal samples produce a large reconstruction error. In contrast, the traditional memory module can successfully reconstruct the class `3' and `4' samples with a small reconstruction error. Therefore, it is difficult to solve the anomaly detection problem by using the traditional memory module to learn the logical pixels, while the proposed PMB module can ensure the storage of normal feature information and use it for normal feature reconstruction. It can be found the similar results in the MVTec AD dataset, the PMB module has a large reconstruction error for abnormal samples.

As shown in Figure \ref{fig:moreexample}, we show the reconstruction results of the PMB module for more abnormal and normal samples. In the `Normal sample' on the left side of Figure \ref{fig:moreexample}, the reconstruction error is small. In the 'Abnormal Sample' on the right, the reconstruction error of the anomalous region shown in the rectangular box is large, while the rest of the errors are small. It indicates that the PMB module does not weaken the reconstruction capability of AE, but makes it impossible for anomalous features to successfully address the abnormal feature output, while normal features can successfully address the normal feature output.

\vspace{1.5em}

\begin{figure}
	\centering
	\includegraphics[width=1\linewidth]{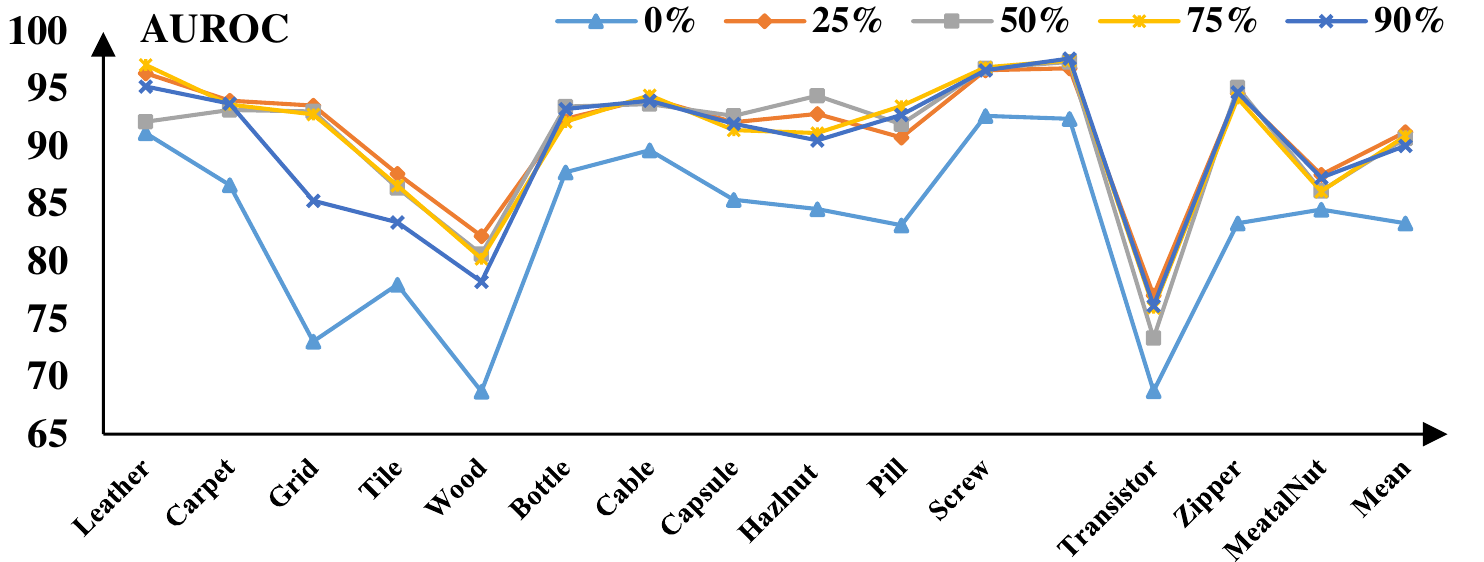}
	\caption{The influence of different values of $\delta_H$ for anomaly localization.}
	\label{fig:res-influ}
\end{figure}

\begin{table}
	\centering
	\caption{The influence of different values of $\delta_H$ for anomaly detection.}
		\renewcommand{\arraystretch}{1.5}
	\setlength{\tabcolsep}{3mm}{
		\begin{tabular}{c|ccccc} \hline
			$\delta_H$&0\%&25\%&50\%&75\%&90\%\\ \hline
			AUROC&88.81&90.1&90.56&91.05&89.96\\

			\hline
	\end{tabular}}
	
	\label{his_d}
\end{table}

\par
\noindent
\textbf{Compare with different AE methods and memory module methods. } To show the advantages of the proposed method over the reconstruction-based approach, the comparison results of different AE structures and memory modules are shown in Table \ref{tab:AEcom}. The different baselines are set up as follows :

\begin{itemize}

	\item[$\bullet$]	\emph{AE-SSIM: }Vanilla AE structure using SSIM loss function.
	\item[$\bullet$]	\emph{AE-L$_2$: }Vanilla AE structure using L$_2$ loss function.
	\item[$\bullet$]	\emph{AE+skip: }Vanilla AE structure using L$_2$ loss function with skip connections.
	\item[$\bullet$]	\emph{MemAE: }Vanilla AE structure using L$_2$ loss function with traditional memory module (Figure \ref{fig:tuhw1} (1)).
	\item[$\bullet$]	\emph{DAAD: }Vanilla AE is equipped with
multi-scale block-wise memory module (Figure \ref{fig:tuhw1} (2)). 
	\item[$\bullet$]	\emph{DAAD+ : }DAAD+ is
DAAD complemented with the adversarially learned representation.
	\item[$\bullet$]	\emph{OURS W/O H: }Vanilla AE is equipped with PMB module.
	\item[$\bullet$]	\emph{OURS: }Vanilla AE is equipped with PMB module and  parameter-free Histogram Error Estimation module.
\end{itemize}
The experimental results show that the proposed method improves 27\%-20\% compared to the vanilla AE method, and improves compared to both MemAE and DAAD methods. These indicate the effectiveness of the novel query generation method of the PMB module, which allows the memory module to store normal features. As in the `WOOD' class, AE+skip performance can reach 97.7\%, while MemAE, DAAD, DAAD+ all reduce the performance, and the proposed method further improves the anomaly detection performance. Comparing the results of MemAE, DAAD, and OURS W/O H, these show that the proposed query (Figure \ref{fig:tuhw1} (3)) is the most effective among the three query generation methods shown in Figure \ref{fig:tuhw1}.

\vspace{1.5em}
\begin{figure}
	\centering
	\includegraphics[width=1\linewidth]{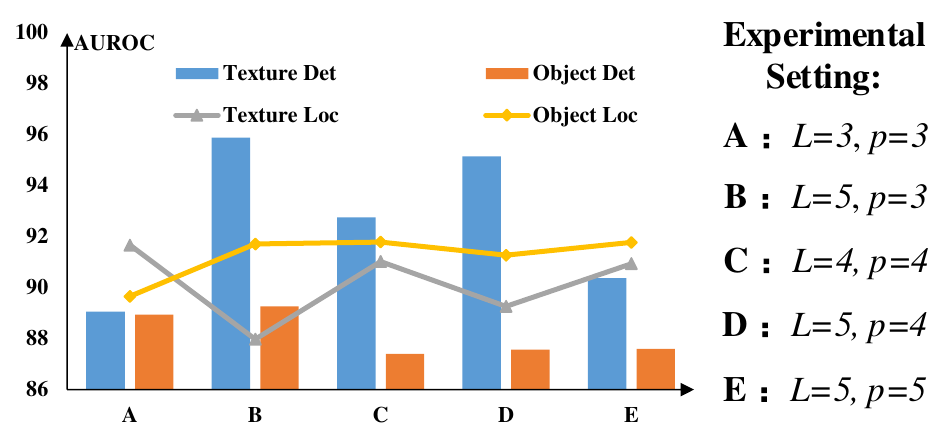}
	\caption{The influences of the combination of skip connection and PMB  on anomaly detection and anomaly localization. 'Det' denotes anomaly detection and 'Loc' denotes anomaly localization.}
	\label{fig:lpmb}
\end{figure}

\par
\noindent
\textbf{Impact of Histogram Error Estimation module.} 
To verify the effect of the proposed Histogram Error Estimation module for anomaly detection and localization, experiments are conducted by varying $\delta_H$ within [0\%, 25\%, 50\%, 75\%, 90\%]. The anomaly detection results are shown in Table \ref{his_d}. When $\delta_H =0\%$, which means that the Histogram Error Estimation module is not exploited, the detection performance dramatically decreases. As $\delta_H$ is increased to 75\%, the performance of anomaly detection is improved and is 2\% higher than the performance without the error estimation module. When $\delta_H=90\%$, the result of anomaly detection decreases to 89.96\%. The overall results in Table \ref{his_d} show that the Histogram Error Estimation module can well estimate the unfavorable reconstruction error caused by AE and effectively improve the performance of anomaly detection. However, when $\delta_H$ is close to 90\%, the performance of anomaly detection is decreased. The reason is anomaly region errors affecting the estimation of $\mu$, resulting in a large estimation bias in $\mu$. For anomaly localization, after eliminating the reconstruction error caused by AE, $A^\prime$ highlights the favorable reconstruction error of the anomaly regions caused by the PMB module, which significantly improves the anomaly localization. From Figure \ref{fig:res-influ}, it can be found that the localization performance of all categories is greatly improved when the proposed histogram error estimation module is introduced. By comprehensively considering the performance of anomaly localization and detection, $\delta _H$ is set to 50\% in experiments.

\vspace{1.5em}

\begin{table}
	\centering
	\caption{The influence of different experimental settings of $k$ values for MVTec AD.}
	\renewcommand{\arraystretch}{1.5}
	\setlength{\tabcolsep}{2.5mm}{
		\begin{tabular}{c|ccccccc}\hline
			MVTec AD&k0&k1&k2&k3&k4&k5&k6\\ \hline
			Det&85.6&83.9&89.6&90.7&88.8&90.4&91.5\\
			Loc&93.0&91.1&91.1&90.8&89.7&89.1&89.5

			\\ 	\hline
	\end{tabular}}
	
	\label{tab:k}
\end{table}
\begin{table}[]
	\centering
	\caption{The influence of different $k$ experimental settings on anomaly detection in MNIST and Retinal-OCT datasets.}
	\renewcommand{\arraystretch}{1.5}
	\setlength{\tabcolsep}{2.5mm}{
		\begin{tabular}{c|cccc}\hline
			\multicolumn{1}{c|}{Det} & \multicolumn{1}{c}{k=(1,1,1)} & \multicolumn{1}{c}{k=(2,2,2)} & \multicolumn{1}{c}{k=(4,4,4)} & \multicolumn{1}{c}{k=(8,8,1)} \\\hline
			MNIST   & 98.1               & 96.5               & 97.6               & 95.8         \\
			
			Retinal-OCT&96.8&98.0&96.8&-\\ \hline              
	\end{tabular}}
	\label{tab7}
\end{table}
\par
\noindent
\textbf{Impacts of $L$ and $p$.}
In the proposed method, the combination strategy of the proposed PMB module and the skip connection are utilized. Experiments are conducted on the MVTec AD dataset to study the sensitiveness of $L$ and $p$. The results are shown in Figure \ref{fig:lpmb}. From the results by varying $p$ from $3$ to $5$ with $L=5$, it can be seen that the alternate utilization of the skip connection and PMB can improve the detection performance, especially for texture images. Compared with the setting `E' which does not use the skip connection, the performance by using settings `B' and `D' is improved by 6\%. Because the detailed features provided by the PMB module guarantee the reconstruction ability of samples while the skip connection provides high-level information to keep the category consistent. For anomaly localization, it is found that the larger the $p$, that is, the more PMBs are introduced, the better the localization performance. In particular, the performance changes more significantly in the texture images, indicating that multiple PMBs can learn detailed normal features more effectively for subtle anomaly localization.


\vspace{1.5em}
\par
\noindent
\textbf{Impact of $k$.}
To evaluate the effect of the number of partitions $k$ on anomaly detection, experiments with different settings were conducted on three datasets. In the MVTec AD dataset, the partitions were set to k0: $(2, 2, 2, 2)$, k1: $(8, 8, 8, 8 )$, k2: $(16, 16, 16, 8 )$, k3: $(32, 16, 16, 8 )$, k4: $(32, 32, 16, 8 )$, k5: $(32, 16, 16, 16 )$, k6: 
$(32, 32, 16, 16 )$. $\delta_H$ is taken as 50\%. The results of anomaly detection and localization on MVTec AD are shown in Table \ref{tab:k}. The results of anomaly detection on the remaining datasets are shown in Table \ref{tab7}. For the anomaly detection task, on the MVTec AD dataset,  The larger $k$ the better the anomaly detection performance. Conversely, the smaller $k$ for MNIST and Retinal-OCT datasets, the better the detection performance. The reason is that the MVTec dataset has a large resolution and carries a large amount of information. Thus multiple storage units are adapted to store the detailed features to ensure the quality of normal samples reconstruction. While MNIST and Retinal-OCT have low resolution and simple semantic information, too large $k$ will make the anomaly features easy to generalize. Therefore, a smaller $k$ is adopted. For the anomaly localization task, the experimental results show that the smaller the $k$ value, the better the localization performance, while the larger the $k$ value, the smaller the impact on localization. The reason is that the expressiveness is relatively weaker when $k$ becomes smaller, and the anomalous features are less likely to be expressed as normal features. Therefore, after eliminating the errors with the histogram error estimation module, the anomaly localization performance is excellent. It is worth noting that the values of $k$ for anomaly localization and anomaly detection are not contradictory. Larger $k$ values indicate a larger number of partitions, more detailed features can be learned, and the overall reconstruction error of normal samples is smaller. Therefore, a large $k$ value is more suitable for anomaly detection. A smaller $k$ value learns limited normal features and better amplifies the error after reconstruction of normal and abnormal pixels. Due to the Histogram Error Estimation module, excellent localization results can still be achieved after adaptive elimination of the unfavorable error.

\vspace{1.5em}

\section{Conclusion}
\par
This paper proposes a new unsupervised visual anomaly detection method by jointly exploring AE and a novel memory module. To address the problem of the existing memory module, a new PMB module with a novel query generation method can learn and store the features of normal samples. With the successful reconstruction of normal features, it makes abnormal features only fit the normal features stored in PMB, which results in the reconstruction error of abnormal regions being larger. Furthermore, to eliminate the cumulative reconstruction error caused by AE, a novel Histogram Error Estimation module is proposed by exploring the reconstruction errors of normal regions. The detection and localization performance is significantly improved. Finally, we explored the optimal combination of memory module and skip connections. Extensive experiments are conducted on three widely-used datasets to verify the effectiveness of the proposed method for visual anomaly detection. In the future,  we will explore a more suitable model to estimate the error caused by AE and will explore the generalizability of the proposed PMB module to other computer vision tasks.

{
	\bibliographystyle{IEEEtran}
	\bibliography{reference}
}

\begin{IEEEbiography}[{\includegraphics[width=1in,height=1.25in,clip,keepaspectratio]{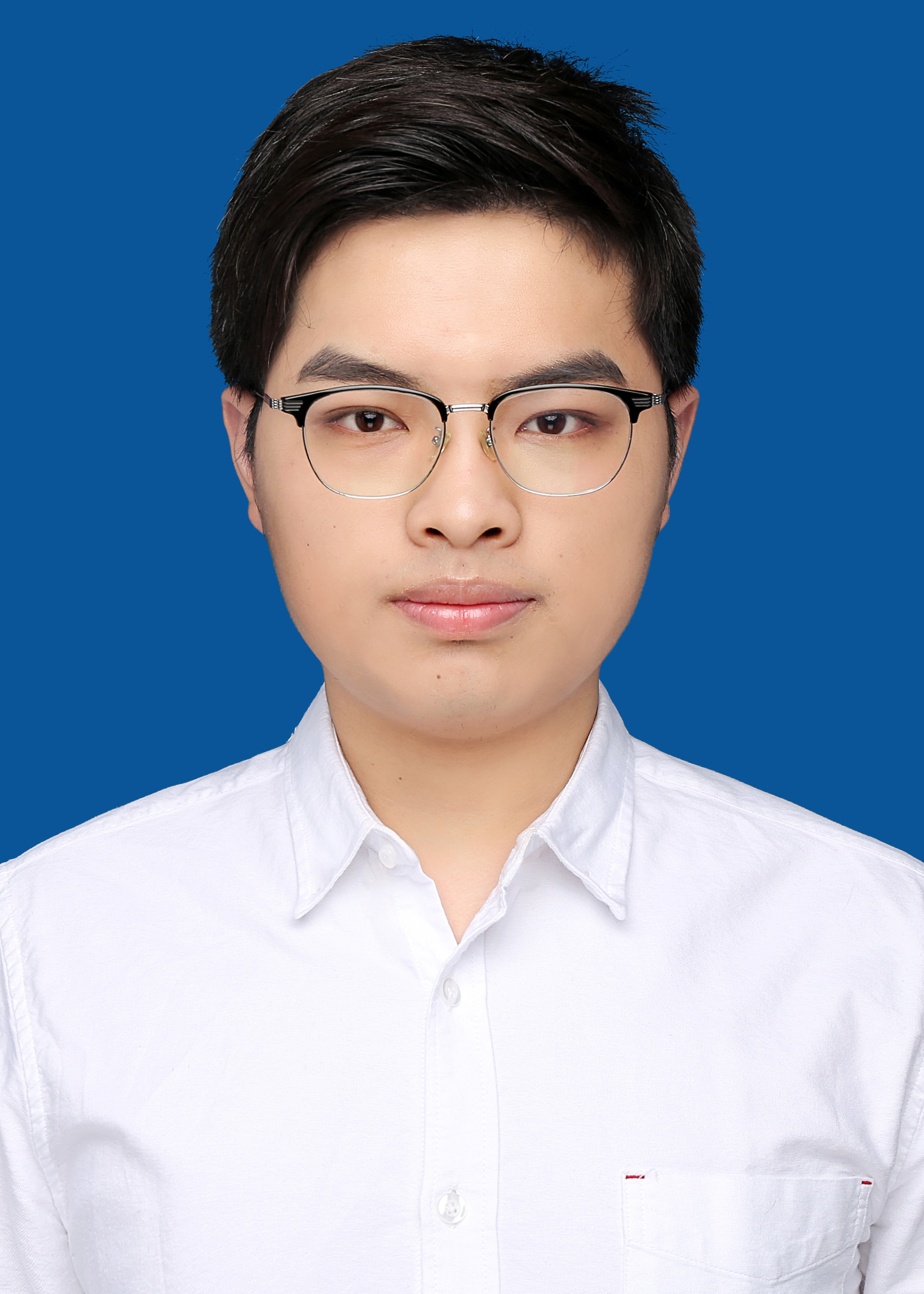}}]{Peng Xing} is currently pursuing the master’s
	degree with the School of Computer Science and
	Engineering, Nanjing University of Science and
	Technology. His current
	research interests include anomaly detection and
	unsupervised deep learning.
\end{IEEEbiography}

\vspace{11pt}
\begin{IEEEbiography}[{\includegraphics[width=1in,height=1.25in,clip,keepaspectratio]{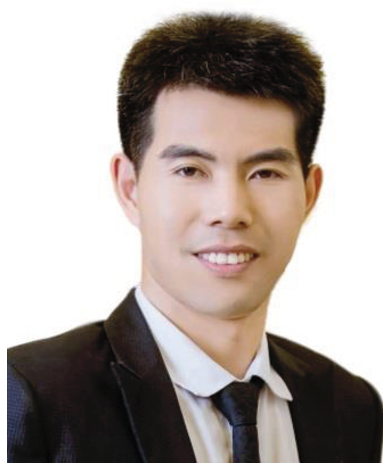}}]{Zechao Li} is currently a Professor at the Nanjing University of Science and Technology. He received his Ph.D degree from National Laboratory of Pattern Recognition, Institute of Automation, Chinese Academy of Sciences in 2013, and his B.E. degree from the University of Science and Technology of China in 2008. His research interests include big media analysis, computer vision, etc. He was a recipient of the best paper award in ACM Multimedia Asia 2020, and the best student paper award in ICIMCS 2018. He serves as an Associate Editor for IEEE \scshape Transactions on Neural Networks and Learning Systems.
\end{IEEEbiography}

\vfill

\end{document}